\documentclass[conference]{IEEEtran}
\IEEEoverridecommandlockouts
\usepackage{cite}
\usepackage{amsmath,amssymb,amsfonts}
\usepackage{algorithmic}
\usepackage{graphicx}
\usepackage{textcomp}
\usepackage{xcolor}
\usepackage{multirow}
\usepackage{soul}
\usepackage{xurl}
\usepackage{hyperref}

\def\BibTeX{{\rm B\kern-.05em{\sc i\kern-.025em b}\kern-.08em
    T\kern-.1667em\lower.7ex\hbox{E}\kern-.125emX}}
\begin{document}

	\begin{titlepage}
		\begin{center}	
		
		\huge
		\textbf{ML-based Short Physical Performance Battery future score prediction based on questionnaire data		
		}

		\vspace{0.5cm}
		\LARGE
		Accepted version
		
		\vspace{1.5cm}
		
		\text{Marcin Kolakowski $^1$, Seif Ben Bader $^2$}
		
		\vspace{.5cm}
		\Large
		$^1$ Institute of Radioelectronics and Multimedia Technology,\\Warsaw University of Technology,
		Warsaw, Poland,\\
		contact: marcin.kolakowski@pw.edu.pl
		
		\vspace{0.5cm}
		
		$^2$ Octilium, Lugano, Switzerland

		\vspace{2cm}

	\end{center}
	
	\Large
	\noindent
	\textbf{Originally presented at:}
	
	\noindent
	2024 32nd Telecommunication Forum (TELFOR), Belgrade, Serbia
	
	\vspace{.5cm}
	\noindent
	\textbf{Please cite this manuscript as:}

	\noindent
M. Kolakowski and S. B. Bader, "ML-based Short Physical Performance Battery future score prediction based on questionnaire data," 2024 32nd Telecommunications Forum (TELFOR), Belgrade, Serbia, 2024, pp. 1-4, doi: 10.1109/TELFOR63250.2024.10819122.
	
	\vspace{.5cm}
	\noindent
	\textbf{Full version available at:}
	
	\noindent
	\url{https://doi.org/10.1109/TELFOR63250.2024.10819122}

			\vspace{.5cm}
			\noindent
			\textbf{Additional information:}
			
			\noindent
			Continuation of the study (prediction of performance in Intrinsic Capacity domains) is presented in:
			
			\vspace{0.5cm}
			
			\noindent
			Kolakowski, M.; Lupica, A.; Ben Bader, S.; Djaja-Josko, V.; Kolakowski, J.; Cichocki, J.; Ayadi, J.; Gilardi, L.; Consoli, A.; Mocanu, I.G.; et al. CAREUP: An Integrated Care Platform with Intrinsic Capacity Monitoring and Prediction Capabilities. Sensors 2025, 25, 916. \url{https://doi.org/10.3390/s25030916} 
	
	\vfill
	
	\large
	\noindent
© 2024 IEEE.  Personal use of this material is permitted.  Permission from IEEE must be obtained for all other uses, in any current or future media, including reprinting/republishing this material for advertising or promotional purposes, creating new collective works, for resale or redistribution to servers or lists, or reuse of any copyrighted component of this work in other works.
\end{titlepage}

\title{ML-based Short Physical Performance Battery\\ future score prediction based on questionnaire data \\
\thanks{
\noindent This research was funded by the Polish National Centre for Research and Development, grant number  AAL/AALCall2021/52/CAREUP/2022.
\newline
The research uses data from the English Longitudal Study of Ageing (ELSA).	ELSA is funded by the National Institute on Aging (R01AG017644), and by UK Government Departments coordinated by the National Institute for Health and Care Research (NIHR).
}
}

\author{

\IEEEauthorblockN{Marcin Kolakowski}
\IEEEauthorblockA{\textit{Institute of Radioelectronics and Multimedia Technology} \\
\textit{Warsaw University of Technology}\\
Warsaw, Poland \\
marcin.kolakowski@pw.edu.pl}

\and

\IEEEauthorblockN{Seif Ben Bader}
\IEEEauthorblockA{
\textit{Octilium}\\
Lugano, Switzerland \\
seif.benbader@octilium.ch}

}

\maketitle

\begin{abstract}
Effective slowing down of older adults' physical capacity deterioration requires intervention as soon as the first symptoms surface. In this paper, we analyze the possibility of predicting the Short Physical Performance Battery (SPPB) score at a four-year horizon based on questionnaire data. The ML algorithms tested included Random Forest, XGBoost, Linear Regression, Dense Neural Networks, and the TabNet architecture. The best results were achieved for the XGBoost (mean absolute error of 0.79 points). Based on the Shapley values analysis, we selected smaller subsets of features (from 10 to 20) and retrained the XGBoost regressor, achieving a mean absolute error of 0.82.   
\end{abstract}


\section{Introduction}
One of the human capabilities that deteriorates with age is physical mobility. The physical capacity deterioration usually manifests as slowed-down movement accompanied by pain, stiff joints, and balance problems.
Since these symptoms significantly affect daily tasks and activities, it is essential to prioritize slowing the progression of physical decline.

Considering the well-established principle that prevention is better than cure, it is vital to take proactive measures at the earliest opportunity, ideally before any signs of deterioration become noticeable. However, screening for future decline is not easy and straightforward as it may require a complex analysis of the patient's health and everyday functioning \cite{prediction_complex}. Those investigations may necessitate conducting additional sets of tests and clinical investigations, which can further increase the workload on clinicians. A potential solution to streamline this process is using machine learning (ML) algorithms, which can be trained on relevant patient data to enhance prediction and early detection. In the literature, several physical capacity prediction methods have been described.

In \cite{imu}, the current user's physical capacity is assessed by predicting scores of two tests, Timed up and Go and Short Physical Performance Battery (SPPB), based on signals registered with an inertial measurement unit worn by the patient. The neural network used for this purpose solves a classification problem. For SPPB score prediction, the accuracy is about 92\%.

A similar approach is presented in \cite{insole_frailty}, where data from an insole sensor is used to predict whether the patient suffers from physical frailty. The study analyzes three classification methods: Random Forests, Linear Regression with cutoffs, and kNN classifier. The highest achieved accuracy was about 72\%.

The studies presented in \cite{socioeconomic_factors, physical_factors} use a sensor-less approach for predicting physical decline associated with sarcopenia. Both studies analyze different classification methods. In both, the classification accuracy does not exceed 85\%.

 In \cite{mobility_activities} a method for predicting incoming problems with performing physical activities (e.g. carrying heavy objects) is presented. The ML prediction based on questionnaire data yields sensitivity from 0.59 to 0.70 depending on activity type.

The work presented in \cite{chinese} aims to classify the patients belonging to two groups those whose cognitive and physical functions are on decline and those whose performance remains the same. The study tests a variety of classifiers arriving at 78\% balanced accuracy.

A similar approach is presented in \cite{knee} where a Random Forest Classifier is used to detect patients, who are at risk of knee capacity decline. The used features including demographics and accelerometer-derived activity measures allowed to achieve accuracy of about 76\%.

This paper introduces a novel approach. Instead of evaluating the patient's current physical performance, the study focuses on predicting future performance in four years. Specifically, we explore the feasibility of forecasting the Short Physical Performance Battery (SPPB) score four years in advance, using questionnaire data related to demographics, health status, and socio-economic factors. To achieve this, we train multiple machine learning regressors on data from the English Longitudinal Study of Aging (ELSA)\cite{elsa}.

\section{Short Physical Performance Battery}
There are several mobility assessment scales intended for elderly screening. The instruments usually target similar mobility aspects like balance, walking speed, and changing body position. The differences typically concern the measurement procedure and the number of tests performed. One of the most renowned instruments is the Short Physical Performance Battery (SPPB) \cite{sppb}. The SPPB assesses the patient's balance, walking speed, and ability to stand up from a chair. It is usually used to screen for several conditions, including frailty \cite{insole_frailty}.

The SPPB (Short Physical Performance Battery) test procedure is relatively straightforward and consists of three timed tests. The times recorded during these tests are converted into scores based on specific thresholds \cite{sppb}. The first part assesses the patient's balance through three timed exercises in which the individual stands in different positions:

\begin{itemize}
    \item feet side by side (SBS) - 0 or 1 point,
    \item one foot slightly in front of the other (semi-tandem) - 0 or 1 point,
    \item one foot placed directly in front of the other (full-tandem) - from 0 to 2 points.
\end{itemize}

The second part assesses gait speed, where the patient is asked to walk four meters. The third part evaluates the patient's ability to change positions by instructing them to rise from a seated position five times without using their arms for support. Depending on time needed for completion gait and chair stand tests are scored from 0 to 4 points each. 

The total score, the sum of the individual test scores, reflects the person's overall physical ability. Scores of 10 or higher indicate good physical condition, while scores between 4 and 9 suggest reduced physical performance but with the potential for recovery. Scores below 4 indicate very poor physical performance and a higher mortality risk. Research has demonstrated that the SPPB score tends to decline with age, with the rate of decline accelerating after the age of 80\cite{declineSPPB}.

\section{Dataset}
\subsection{Data source}
\begin{figure}[bp]
	\centerline{\includegraphics[width=0.9\linewidth]{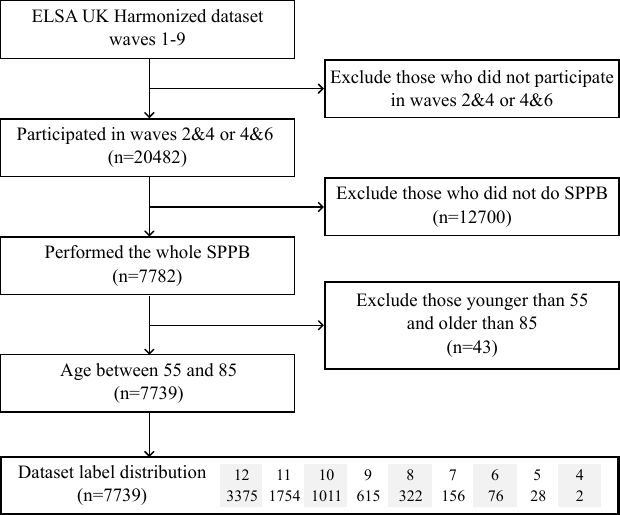}}
	\caption{Data selection process}
	\label{fig:selection}
\end{figure}

We used the harmonized dataset of the English Longitudinal Study on Aging (ELSA) \cite{elsa}. The dataset contains data gathered in the United Kingdom in nine waves conducted from 2002 to 2019. It includes data from close to twenty thousand individuals who took part in at least one wave of the study. The data gathered in each wave contains information about various aspects of the participant's life, including demographics,  health, family, finances, employment, and performance in several areas, such as cognition or physical capacity. The process of data selection is illustrated in Fig. \ref{fig:selection}.

Physical capacity in ELSA is assessed using tests based on SPPB. Selected participants were asked to perform balance, walking, and chair stand tests. As the taking appropriate measurements requires professional personnel assistance, they were conducted only in selected waves: 2,4,6 spaced by four years. Based on such data, future physical performance predictions can be made at 4- and 8-year horizons only.

The dataset was constructed by taking the person's answers from waves 2 and 4 and combining them with SPPB scores calculated based on subsequent waves 4 and 6 results, respectively. As in the ELSA study, the walking distance (8 ft) is different from the standard one (4 m),  the scoring thresholds were adjusted proportionally.

The individuals younger than 55 and older than 85 were excluded, which resulted in a dataset of 7739 samples. The dataset was not balanced, as the participants with high future SPPB scores outnumbered those with mobility problems.

\subsection{Feature selection}
In each wave of the ELSA study, the participants are asked about a thousand questions concerning various aspects of their lives, from health state to work and living conditions. In the study, we used a subset of data concerning health state, physical performance, demographics, and capability to do basic daily activities. The list of the parameters used is presented in Table \ref{tab:features}.

The dataset includes both continuous and categorical variables. In most cases, the discrete responses are binary (e.g., yes/no), or the categories are encoded in a monotonic manner. For instance, for self-rated health, lower values correspond to better health, while higher values indicate worsening health. The only variable that was one-hot encoded was marital status.

The dataset also included additional derived features, such as the partial SPPB scores for each test and the total SPPB score.

\begin{table*}[t]
\caption{ELSA UK variables used for SPPB score prediction}
\centering
\def\arraystretch{1.3}%
\begin{tabular}{p{0.14\linewidth}  p{0.84\linewidth}}
	\hline
	category & variables \\ 
	\hline\hline
	demographics & age, marital status, gender, education, family information: parents age/age of death, no. of household members \\
	\hline
	health state & self-rated health, dental health , ever diagnosed with: high blood pressure, diabetes, cancer, lung disease, heart problems, stroke, arthritis, asthma, high cholesterol, cataracts, Parkinson's, hip fracture, osteoporosis, psychical problems, experiences: shortness of breath/persistent wheezing, problems with pain, leg pain  \\
	\hline
	medical procedures & undergone any joint or hip replacements \\
	\hline
	recent medical history & whether experienced angina, heart attack, fractured hip or psychological problems in last 2 years\\
	\hline
	physical capabilities & whether it is difficult to: walk 100 yards, sit for 2 hours, carry 10 lbs, pick up a 5p coin, extend arms, push large objects \\
	\hline
	sensory capabilities & self rated eyesight (distance and near) and hearing \\
	\hline
	cognitive functions & CESD score, whether experienced memory problems \\
	\hline
	falls \& outcomes & whether has fallen down in last 2 years, no. of falls, whether sustained injuries, whether uses fall alarm aids,.\\
	\hline
	physical performance & SPPB tests time measurements in seconds (3 balance tests, walking speed, chair stands), corresponding scores, a total SPPB score, grip strength in kg \\
	\hline
	daily functioning & whether health limits work, Activities of Daily Living (all 6), Instrumental Activities of Daily Living (all excluding communication and danger recognition)\\
	\hline
	habits & whether smokes/smoked, no. of cigarettes/day, whether drinks alcohol, no. of drinks/week, participation in social events, working status, frequency of vigorous, moderate and light physical activity,  \\
	\hline
	physical measures & height, BMI, systolic and diastolic pressure \\
	\hline
\end{tabular}
\label{tab:features}
\end{table*}

\section{Experiments}

\subsection{Methods}
We have tested several ML models that were proven to work well with this kind of tabular data: Random Forest Regression, XGBoost, Linear Regression, and two types of neural networks: dense and TabNet. We have tested the models with different hyperparameters listed in Table \ref{tab:parameters}.

Random Forest Regression and XGBoost are ensemble methods in which the output of several smaller models is combined to ensure better model generalization. Both of the methods are tree-based. The significant difference is that in Random Forests the trees are trained separately whether XGBoost uses the input from the previously trained trees to improve accuracy. Both methods were tested for different numbers of trees and maximum depths.

Linear regression fits a linear model to the supplied dataset. In the case of our study, a default implementation was used.

In the study, we have trained dense neural networks of different numbers of layers and neurons. Each dense layer except the last one was followed by batch normalization to improve generalization.

TabNet \cite{tabnet} is an architecture  for working with tabular data. In TabNet architecture, the data are processed in steps in which the reasoning is performed based on features selected using the attentive mechanism. The parameters tested in the study were the number of steps and the gamma value, which determines how independent the reasoning is between the steps.

\begin{table}[b]
	\caption{Tested algorithm parameters}
	\centering
	\def\arraystretch{1.3}%
	\begin{tabular}{l  c c}
		\hline
		model & parameter & parameter grid \\ 
		\hline
		\hline
		Random Forest & trees no. & 10, 50, 100, 200, 300\\
		&max depth&  2, 8, 16, 32, 64, None\\
		\hline
		XGBoost & trees no.  & 10, 50, 100, 200, 300\\
		&max depth&  2, 8, 16, 32, 64, None\\
		\hline
		Linear Regression & - & -\\
		\hline
		TabNet & max steps & 2, 4, 6,  8\\
		& gamma & 1.2,  1.8,  2.4,  3\\
		\hline
		Dense Neural Network & layers & 2, 3, 4, 5\\
		& neurons/layer & 8, 16, 32, 64, 128 \\
		\hline
	\end{tabular}
	\label{tab:parameters}
\end{table}

\subsection{Data preprocessing}
Before training the algorithm, the dataset underwent preprocessing. Missing values were imputed using the KNN imputer provided by the sklearn library, and the variables were scaled to a 0–1 range.

To address dataset imbalance, we conducted tests where examples from underrepresented classes were either oversampled or augmented with additional noise. However, these methods did not result in a significant performance improvement. Therefore, the results presented in the subsequent sections are based on the original dataset without further augmentation.

\section{Results}

The accuracy of the models was evaluated using a 10-fold cross-validation scheme. The results varied depending on the chosen hyperparameters. The accuracy metrics, including Mean Absolute Error (MAE) and Mean Squared Error (MSE), along with the corresponding hyperparameters, are summarized in Table \ref{tab:accuracy}.

\begin{table}[b]
	\caption{Future SPPB score prediction accuracy of the tested methods}
	\centering
	\def\arraystretch{1.3}%
	\begin{tabular}{l  c c c}
		\hline
		model & MAE & MSE  & parameters\\ 
		\hline
		Random Forest &0.80014 & 1.14870 & trees:300, depth:16\\
		XGBoost &   0.79259 & 1.15076& trees:100, depth:2 \\
		Linear Regression& 0.80151 &1.13315  & -\\
		TabNet & 0.80643 &	1.16067 & steps: 6 gamma: 3\\
		Dense Neural Network & 0.87822 & 1.44207 & layers size: 8 16 8\\
		\hline
	\end{tabular}
	\label{tab:accuracy}
\end{table}

The highest accuracy was achieved with the XGBoost model, which produced an MAE of approximately 0.79 SPPB points. However, the other models, except for the dense neural networks, showed comparable accuracy. This suggests that for the problem at hand—predicting the SPPB score over four years based on selected features—the accuracy may be approaching its theoretical limit. This could be attributed to how the SPPB score is calculated, where even a minor change in the time taken to perform tasks like chair stands or a walking test can reduce the score by an entire point. Despite these limitations, the achieved accuracy provides valuable insights into the likely trend of the SPPB score, allowing patients to take timely actions based on predicted changes.

\subsection{Simplified models}
\begin{figure}[tbp]
	\centerline{\includegraphics[width=\linewidth]{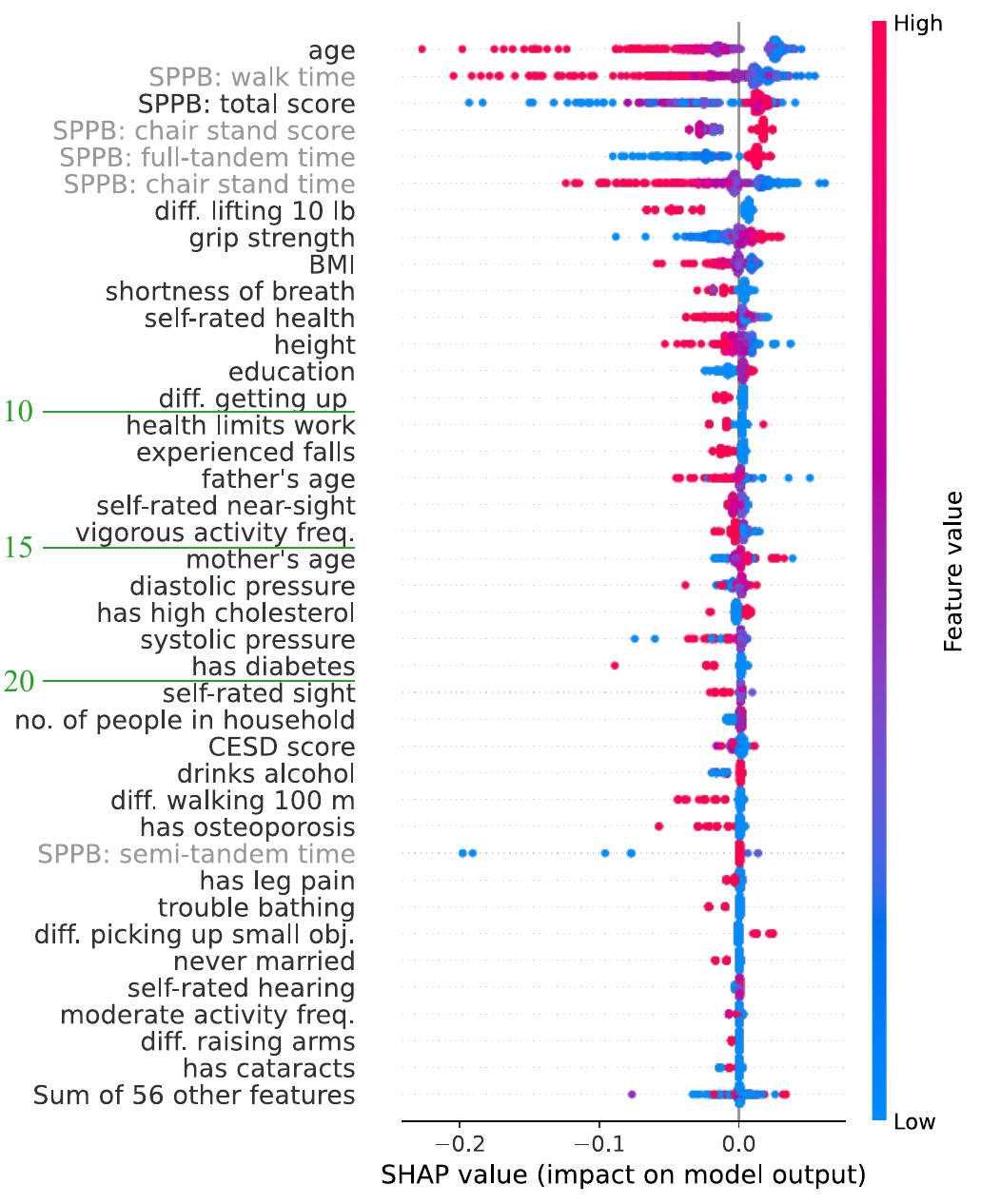}}
	\caption{Beeswarm illustrating the feature importance for the XGBoost model. The 10/15/20 lines mark sets of values used for the simplified models.}
	\label{fig:shapley}
\end{figure}

The importance of features was assessed based on their Shapley values. The values were computed using the SHAP Python package for the XGBoost Regressor. The beeswarm plots for the most salient features are presented in Fig. \ref{fig:shapley}. The most influential features included SPPB-related variables: scores and measured times, a reference point for future predictions. Besides them, the most important were: age, grip strength, BMI, self-rated health, and strength.

Based on the Shapley values analysis, the XGBRegressor (no. of trees 100, depth 2) was trained based on a smaller number of the most important features. The used features did not include SPPB-related scores and times apart from the current total SPPB score. The model was trained based on the 10, 15, and 20 most salient features, and MAEs of 0.8175, 0.8150, and 0.8162 were achieved, which is a small difference compared to 0.7925 obtained for all 95 features. That means the number of questions used for predicting physical decline may be reduced.

\section{Conclusions}
The paper presents the results of the study on future SPPB score prediction based on questionnaire data. In the study, we analyzed the effectiveness of several regression methods on a dataset from the ELSA UK study. The best accuracy was achieved for the XGBoost regressor, and the MAE value was about 0.79 SPPB points. We selected smaller subsets of 10, 15, and 20 features based on the Shapley values. The accuracy obtained for the simpler models was at similar levels.

Given that the tested methods yielded similar accuracy, the way to improve the prediction accuracy may lie in rephrasing the problem: changing the prediction horizon or choosing a different dataset. Adding more relevant features like gait parameters or raw inertial signals registered during daily activities might improve efficiency.

\end{document}